\documentclass{article}

\PassOptionsToPackage{numbers, compress}{natbib}

\usepackage[preprint]{neurips_2019}
\usepackage{listings}

\usepackage[utf8]{inputenc} 
\usepackage[T1]{fontenc}    
\usepackage{hyperref}       
\usepackage{url}            
\usepackage{booktabs}       
\usepackage{amsfonts}       
\usepackage{nicefrac}       
\usepackage{microtype}      
\usepackage{algorithmic}
\usepackage{algorithm}
\usepackage{amsmath,amsthm,amssymb,amsfonts}
\usepackage{wrapfig}
\usepackage[pdftex]{graphicx}

\usepackage{ulem}

\title{NeMo: a toolkit for building AI applications using Neural Modules}

\author{
 \textbf{Oleksii Kuchaiev, Jason Li, Huyen Nguyen, Oleksii Hrinchuk, Ryan Leary, Boris Ginsburg,} \and \textbf{Samuel Kriman, Stanislav Beliaev, Vitaly Lavrukhin, Jack Cook, Patrice Castonguay}, \and \textbf{Mariya Popova, Jocelyn Huang, Jonathan M. Cohen} \\ \\ NVIDIA \\ Santa Clara, CA \\
 \texttt{\{okuchaiev,jasoli,chipn,ohrinchuk,rleary,bginsburg,skriman,stanislavv,} \\
 \texttt{vlavrukhin,jocook,pcastonguay,mpopova,jocelynh,jocohen\}@nvidia.com}
 }
\begin{document}

\maketitle

\begin{abstract}
NeMo (\textbf{Ne}ural \textbf{Mo}dules) is a Python framework-agnostic toolkit for creating AI applications through re-usability, abstraction, and composition. NeMo is built around neural modules, conceptual blocks of neural networks that take \textit{typed} inputs and produce \textit{typed} outputs. Such modules typically represent data layers, encoders, decoders, language models, loss functions, or methods of combining activations. NeMo makes it easy to combine and re-use these building blocks while providing a level of \textit{semantic correctness} checking via its neural type system. The toolkit comes with extendable collections of pre-built modules for automatic speech recognition  and natural language processing. Furthermore, NeMo provides built-in support for distributed training and mixed precision on latest NVIDIA GPUs. NeMo is open-source.\footnote{Available at: \url{https://github.com/NVIDIA/NeMo}}
\end{abstract}

\section{Introduction}
Deep Learning (DL) has made huge progress from academia to industry in the last decade. However, the process for developing, debugging, and deploying DL software is significantly more cumbersome than other complex software systems. The primary abstraction in all DL frameworks is a multidimensional tensor, typically without any dimensional semantics, e.g. whether the first dimension represents the batch size or something else. The lack of semantics and a type system complicates models' re-use and makes it difficult to build DL systems \citep{namedtensor}. It can be challenging to reuse components of a complex DL model across different use cases or developers. The typical approach for reusing and sharing components is based on open-source pre-trained models. Combining and chaining these models together usually requires making changes to the code, which, in turn, requires debugging. This is especially tricky when the models come from different developers or use cases.

Another complicating factor is that model configurations are usually defined via a Python script instead of a data model. This leads to a blurring of lines between what would otherwise be separate concerns -- computational performance, architecture definition, training procedure, visualization and analysis -- all mixed together into the same Python script that is difficult to disentangle, debug, or reuse in other contexts.

All of these challenges -- separation of concerns, system decomposition with well-defined verifiable interfaces, and code re-usability -- are already well-explored in the world of software engineering.  Many of the modern techniques any software developer takes for granted were originally invented in order to address precisely these issues.

We seek to translate common software engineering practices developed to address those issues into the context of developing AI-based applications. Specifically, we focus on the problems of:
\begin{itemize}
  \item \textbf{decomposition} of a complex system into functional blocks with well-defined interfaces;
  \item \textbf{static type checking} to ensure API compliance and to catch type-mismatch bugs;
  \item \textbf{separation of concerns} between model architecture, training procedure, DL framework, optimization algorithm;
  \item \textbf{high performance training} by supporting modern efficient hardware features; and
  \item \textbf{reusable pre-built components} that can be easily combined in novel ways.
\end{itemize}

NeMo consists of: (1) \textbf{NeMo Core}: fundamental building blocks for all neural models and type system and (2) \textbf{NeMo collections}: pre-built neural modules for particular domains such as automatic speech recognition (ASR), and natural language processing (NLP).

\section{Related work}\label{rel_work}
In recent years, there have been a number of high-level toolkits aimed to help users to achieve certain goals easier then by purely using DL frameworks such as TensorFlow \citep{tensorflow2015-whitepaper} and PyTorch \citep{paszke2017automatic}.

These toolkits could be loosely classified into two main groups: (1) higher-level neural network APIs such as Keras\citep{chollet2015keras}, Sonnet \citep{sonnetblog}, PyTorch Ignite \citep{ignite}, PyTorch Lightning \citep{lightning} and (2) configuration-driven tookits such as Tensor2Tensor \citep{tensor2tensor}, Ludwig \citep{ludwig}, OpenSeq2Seq \citep{kuchaiev2018}, FairSeq \citep{ott2019fairseq}, OpenNMT \citep{klein-etal-2017-opennmt}, Seq2Seq \citep{Britz:2017} and many others.

Conceptually, NeMo Core is closer to the first group. It allows users to express models with arbitrary sets of components and hides away details of training and evaluation loops while still retaining a lot of flexibility. NeMo collections, on the other hand, are closer to the second group. They contain common modules that can be re-used in various scenarios. For example, it is straightforward in NeMo to define templates for fixed patterns, such as an encoder-decoder network.

NeMo differs from toolkits in the first group in two ways: (1) NeMo's core abstraction is a neural module rather than a layer or a tensor and (2) NeMo contains a neural type system that performs various semantic checks.

The main difference between NeMo and toolkits in the second group is that NeMo does not impose any particular structure, e.g. many toolkits require models to follow the encoder-decoder-loss structure. NeMo also does not require configuration files to be in any particular format -- users can define models directly using NeMo API. Some toolkits from the second group enforce input-output compatibility between blocks \citep{PyTorchPipe}, but NeMo does this using a consistent, generic, and extensible type system.

Conceptually, NeMo is similar to PyTorchPipe \citep{PyTorchPipe}, which follows the task-oriented approach, but allows for arbitrary, flexible sets of components, and also performs compatibility checks. It is essentially an application framework, while NeMo allows developer to use its underlying type and composition system, but not otherwise adopt any of the NeMo run-time functionality.

\section{NeMo}
The core building block in NeMo is called \textit{Neural Module} (NM). A Neural Nodule represents a \textit{logical} part of a neural network such as a language model, an encoder, a decoder, a data augmentation algorithm, a loss function, or other sets of layers and functions. As the primary abstraction in NeMo, NMs form the basis for describing a model and the process by which that model is trained.  Formally, a Neural Module is a component that computes a set of \textit{typed} outputs given a set of \textit{typed} inputs. Inputs and outputs are collections of multidimensional tensors. In the same way that a programmer in an object-oriented language can choose at what level of granularity to define an object, a NeMo user can choose the level of granularity of a Neural Module. A basic rule is that inputs and outputs should ``make sense'' to expose via an interface.  This suggests that a Neural Module is not typically a single neural network layer, but rather a collection of connected layers that ``do something useful'' such as an encoder, a concatenation operation, a loss function, or a data augmentation.

\begin{wrapfigure}{r}{0.4\textwidth}
\includegraphics[width=0.4\textwidth]{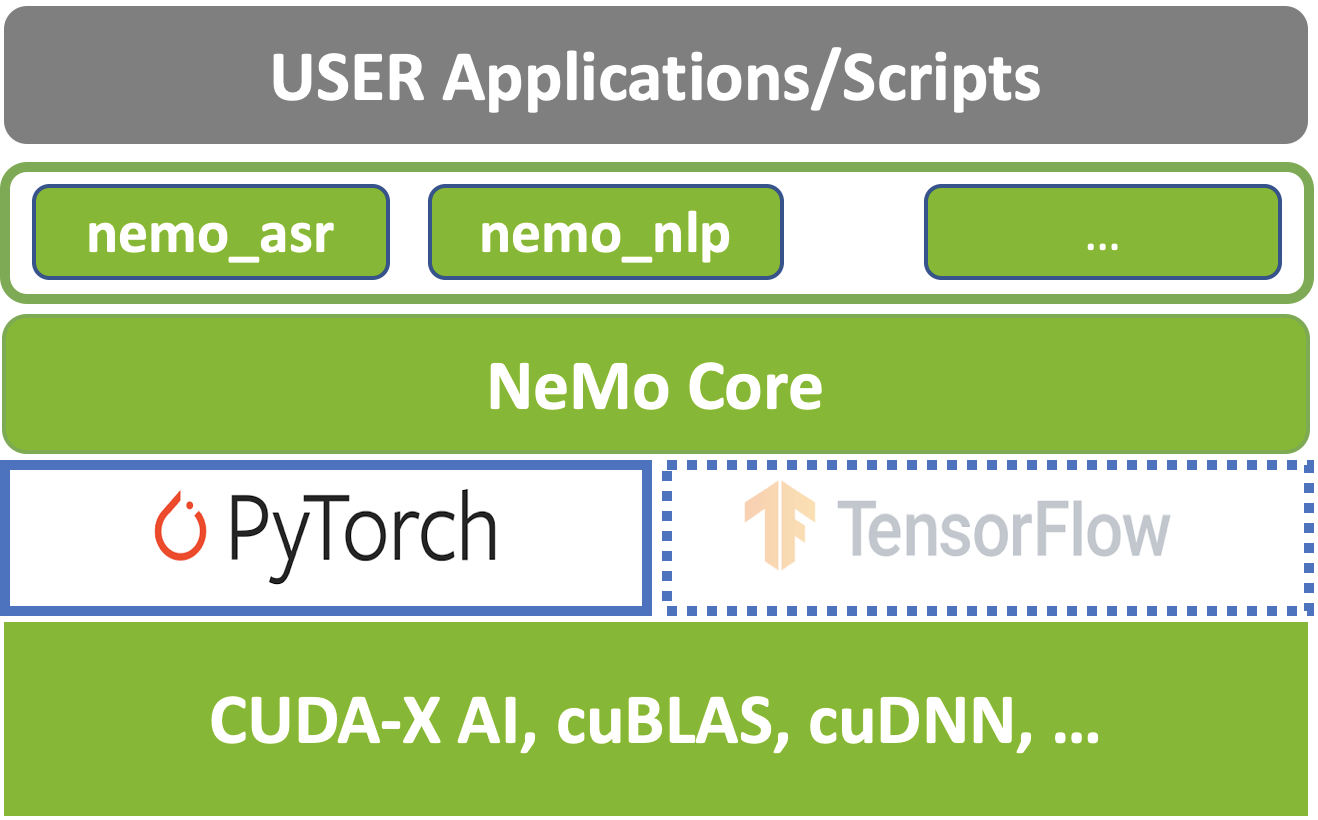}
\caption{NeMo is a framework-agnostic toolkit which serves as abstraction level between application and DL frameworks backend.}
\label{nemo_place}
\end{wrapfigure}

In our implementation, a NM is a Python class that describes its \textit{input ports} and \textit{output ports} using the type system described below. The current implementation relies  on PyTorch, but the abstractions and code does not make any reference to the underlying framework, allowing for applications to be framework-agnostic and for the addition of new backends in the future (see Figure \ref{nemo_place}).

A NM can compute its output port values given provided input port values. For NMs that contain trainable weights, they should also be able to compute the gradient flows. Implementations of the forward and backward passes are provided by the underlying DL framework. One way to think of a NM is that it is able to ``lower'' itself into either another set of NMs (recursively), a set of well-defined neural network layers implemented in PyTorch, or in the case of non-trainable NMs, by directly evaluating the output values given the input values.  

A NM may be parameterized. For example, an image encoder NM can be parameterized by the number of convolutional layers, filter sizes, dropout values, etc. In this way, a NM defines a parametric family of neural networks, where the variation is explicitly determined by parameter values. Parameters are passed to the NM at construction time via named parameters, and the code that defines how lowering occurs only depends on constructor-time values of these parameters\footnote{This is enforced via convention for now.}.

Similar to functional programming languages, the evaluation of a NM's outputs or gradients cannot effect the evaluation of any other NM's inputs or gradients, except via explicitly linked inputs and outputs which follow standard neural network forward and backward propagation rules. This decoupling helps enforce clean and correct decomposition of complex models according to well-defined interfaces because evaluations cannot have side effects. Furthermore, explicitly defined parameters allow experiment tracking and integration with hyperparameter search tools to be greatly simplified. 

This illustrates a major principle of NeMo: \textit{the structure of a neural network and its forward and backward data flows should be determined by values in data structures, not by logic encoded in Python source code}. 

\subsection{NeMo Core}\label{nemo_core}
An application built with NeMo typically consists of 3 required stages and 1 optional stage:
\begin{enumerate}
  \item Instantiate a NeuralFactory object and the necessary NMs. 
  \item Define the activation flow DAG by connecting NMs together.
  \item (optional) Define callbacks for logging, checkpointing, visualization, and evaluation.
  \item Invoke an action such as \texttt{train}, \texttt{eval}, or \texttt{infer}.
\end{enumerate}

NeMo follows a lazy execution model: no computation is done until an action is called. During the definition of the activation flow directed acyclic graph (DAG), NeMo does type checking for the inputs and outputs of connected NMs. This helps catch and debug various errors prior to doing any computations. Once the DAG of modules is defined and action is called, 
NeMo invokes the DL framework, which we call a \textit{backend}.
NeMo is designed to be framework-agnostic, but it currently only supports PyTorch as backend.


Users can create their own NMs by combining existing NMs or providing an implementation in a particular framework. In practice, any PyTorch \texttt{nn.Module} can be easily converted into a NeMo's NM by adding input and output port definitions - i.e. what is expected by and returned from the \texttt{forward} function.

Similar to scikit-learn and Keras, NeMo allows users to create callbacks for routines performed during training such as evaluation, logging, and performance monitoring.

\subsection{Neural types system}\label{types}
NeMo Core defines the interface and functionality of the \texttt{NeuralModule} base class. Each input and output of a NM has a \textit{Neural Type}. Neural Types describe the semantics, axis order, and dimensions of a tensor. The purpose of this type system is to catch semantic and dimensionality errors during model creation and facilitate module re-use. If the output type of one NM output port matches the input type of another NM input port, it is legal to connect these two NMs together, regardless of where they came from or how they are implemented. Because we use static type checking, NeMo can catch semantic and dimensionality errors during the DAG creation stage. 

A Neural Type is a mapping from each tensor axis ID to an \textit{Axis Type}. An Axis Type contains semantic and dimensional information of a tensor's axis. Semantic information is represented with the help of ``semantic tags'' - Python classes related by ``is-a'' kind of inheritance. For example, if module A's output is of the semantic type \texttt{WordEmbedding} while module's B expected input is \texttt{Embedding}, and \texttt{WordEmbedding} is inherited from \texttt{Embedding}, then module B can accept A's output as input but not vice-versa. See Table~\ref{tab:types} for examples.

A \texttt{NeuralType} is constructed from a dictionary, \texttt{axis2type}, which maps an axis index to its \texttt{AxisType}. For example, the input and output ports of a typical ResNet encoder can be described as follows:
\begin{lstlisting}[language=Python, basicstyle=\small]
input_ports = {"x": NeuralType({0: AxisType(BatchTag), 
                                1: AxisType(ChannelTag), 
                                2: AxisType(HeightTag, 224), 
                                3: AxisType(WidthTag, 224)})}
output_ports = {"output": NeuralType({0: AxisType(BatchTag), 
                                      1: AxisType(ImageEmbeddingTag)})}
\end{lstlisting}

NeMo defines binary comparison operation for any pair of \texttt{NeuralType} objects with various comparison results, such as \texttt{SAME}, \texttt{LESS}, \texttt{GREATER}, \texttt{DIM\_INCOMPATIBLE}, \texttt{TRANSPOSE\_SAME}, and \texttt{INCOMPATIBLE}. This type system also allows for non-tensor objects (such as scalars) and the \texttt{root} type which is somewhat analogous to \texttt{void*} in C++: a port of \texttt{root} type can accept any tensor\footnote{In the future, we plan to add a template type system modeled on a simplified version of C++ templates to support type-safe generics for operations such as concatenation.}.

Examples of errors that NeMo's type system can catch at model definition time include: ``\textit{Ranks match but semantics don't}", ``\textit{Concatenating along the semantically wrong dimensions}", and ``\textit{Dimensions mismatch}". For example, consider an encoder-decoder model where the decoder expects input in the form of \texttt{[batch\_dim, time\_dim, channel\_dim]} and the encoder, written by another developer, outputs a \texttt{[time\_dim, batch\_dim, channel\_dim]} tensor. Time and batch dimensions are often dynamic, and if \texttt{channel\_dim} remains constant, standard frameworks will run smoothly but the model will fail to converge, forcing the developer to find and fix this silent error. However, NeMo will throw a semantic type error at the moment these modules are connected. In this case, the result of type comparison operation will be \texttt{TRANSPOSE\_SAME} instead of \texttt{SAME}.\footnote{It is possible to add \textit{implicit casts} where the system automatically inserts simple operations such as transposition to ``fix'' simple type mismatches, similar to how C++ can automatically promote \texttt{int} to \texttt{float}.}.

\begin{table}[!ht]
\centering
\label{tab:types}
\begin{tabular} {l|l|c} 
 \toprule
           Output port type & Input port type & Comparison Result  \\
 \hline 
         \{0: Batch, 1: Channel\} & \{0: Batch, 1: Spectrogram\} & GREATER (INCOMPATIBLE) \\
         \{0: Batch, 1: Spectrogram\} & \{0: Batch, 1: Channel\}  & LESS (COMPATIBLE) \\
         \{0: Batch, 1: Spectrogram\} & \{0: Batch, 1: Encoded\}  & INCOMPATIBLE \\
         \{0: Batch, 1: Spectrogram\} & \{0: Spectrogram, 1: Batch\}  & TRANSPOSE\_SAME \\
        \{0: Batch, 1: Spectrogram:64\} & \{0: Batch, 1: Channel:40\}  & DIM\_INCOMPATIBLE \\
        \{0: Batch, 1: Spectrogram:64\} & \{\}  & SAME \\
         
\bottomrule
\end{tabular}
\caption{Examples of NeMo Neural Types. \{\} denotes ``root'' neural type. These examples assume the following: (1) Spectrogram and Encoded types are inherited from the Channel type (2) Module's A output port is connected to module's B input port.}
\end{table}

\subsection{High performance training}
NeMo is built to take full advantage of the latest DL hardware such as NVIDIA's Volta and Turing GPUs. It automatically supports \textit{mixed precision} training, using \textit{float16} for computationally intensive operations such as matrix multiplies and convolutions while keeping some things in \textit{float32}, and employing dynamic loss scaling\citep{micikevicius2017}.

NeMo also supports gradient accumulation, a technique that accumulates gradients on workers and updates weights only after a certain number of batches have been processed. This allows very large batch simulations on cards with limited RAM and facilitates distributed multi-GPU runs by reducing the amount of inter-worker communication. NeMo also supports multi-GPU and multi-node training using NVIDIA's APEX library\citep{apex}. 

\section{NeMo collections}\label{collections}
A NeMo collection is the DL equivalent of a software collection of related functions. Common NMs for particular domains are pre-built and packaged into ``collections''. Currently, NeMo provides collections for automatic speech recognition (ASR) and natural language processing (NLP), but users can easily add new collections. Some of those NMs may come with pre-trained weights.  A NeMo collection is the DL equivalent of a software library containing a collection of related functions.  In practice, a collection is simply a Python module that defines NM classes, neural types, and associated helper routines. 

\subsection{Automatic Speech Recognition}
\texttt{nemo\_asr} is a collection of neural modules and helper functions that can be used to train and evaluate Automatic Speech Recognition (ASR) models. It currently supports two model types: CTC-based and sequence-to-sequence attention-based.


\subsubsection{CTC-based speech recognition}\label{ctcexample}
As an example, we describe the steps necessary to train a Jasper-like ASR model\citep{li2019} using \texttt{nemo\_asr}.

First, we create a Neural Module Factory object which manages training and instantiation of neural modules. Jasper uses a \texttt{JasperEncoder}, a \texttt{JasperDecoderForCTC}, and a \texttt{CTCLossNM}.  Each of these NMs is passed parameters at construction time - we omit them here to make the code simpler to read.
\begin{lstlisting}[language=Python, basicstyle=\small]
nf = nemo.core.NeuralModuleFactory(...)
data_layer = nemo_asr.AudioToTextDataLayer(...)
jasper_encoder = nemo_asr.JasperEncoder(...)
jasper_decoder = nemo_asr.JasperDecoderForCTC(...)
ctc_loss = nemo_asr.CTCLossNM(...)
\end{lstlisting}

Next, we define the Directed Acyclic Graph (DAG) of how activations flow from output ports to input ports.
\begin{lstlisting}[language=Python, basicstyle=\small]
spec, spec_len, transcript, transcript_len = data_layer()
encoded, encoded_len = jasper_encoder(audio_signal=spec, length=spec_len)
log_probs = jasper_decoder(encoder_output=encoded)
loss = ctc_loss(log_probs=log_probs, targets=transcript,
                input_length=encoded_len, target_length=transcript_len)
\end{lstlisting}

In the first line, the \texttt{data\_layer} produces 4 output ports, which are returned as a tuple. The \texttt{jasper\_encoder} has two named input ports, which are connected to two of the \texttt{data\_layer} output ports.

During the definition of this DAG neural type system checks are performed to ensure the correct usage of various modules together. Finally, once the DAG is described, we should call a neural factory's action, such as \texttt{train} to trigger the training procedure and data flow. 
\begin{lstlisting}[language=Python, basicstyle=\small]
nf.train(tensors_to_optimize=[loss],
         callbacks=[train_cb, saver_cb], ...)
\end{lstlisting}

\subsubsection{Neural modules re-use: attention-based speech recognition}\label{laslike}
As an illustration of model reuse, we show how to build a different ASR model using attention-based sequence learning, but reusing the same data layer and the Jasper encoder. The model we'll build is conceptually similar to LAS \citep{chan2016listen}

\begin{lstlisting}[language=Python, basicstyle=\small]
... //instantiate most modules as before
connector = nemo_asr.JasperRNNConnector(...)
decoder = nemo.common.DecoderRNN(...)
seq_loss = nemo.common.SequenceLoss(...)

spec, spec_len, transcript, transcript_len = data_layer()
encoded, encoded_len = jasper_encoder(audio_signal=spec, length=spec_len)
encoded = connector(tensor=encoded)
log_probs, _ = decoder(targets=transcripts, encoder_outputs=encoded)
train_loss = seq_loss(log_probs=log_probs, targets=transcripts)
\end{lstlisting}
In this example, we switch out \texttt{JasperDecoderForCTC} for a \texttt{DecoderRNN}, and \texttt{CTCLoss} for a \texttt{SequenceLoss} NM. Note, that we use a \texttt{JasperRNNConnector} NM to correct for the dimensionality mismatch between \texttt{JasperEncoder} and \texttt{DecoderRNN}. Notice that \texttt{DecoderRNN} comes from a different collection, and could be first pre-trained as a stand-alone language model.


\subsection{Natural Language Processing}
\texttt{nemo\_nlp} is a collection of neural modules and callback functions which can be used for various NLP-related tasks such as neural machine translation (NMT), language modeling, sentence classification, asr correction, joint intent classification and slot filling. It also supports BERT pre-training and fine-tuning for each task. For BERT, we rely on the implementation from \texttt{pytorch-transformers}\cite{ptransformers}. We plan to extend this collection in future.

\subsubsection{Neural Machine Translation}
Here we show how to build \texttt{Tranformer-BIG} \cite{vaswani2017} using \texttt{nemo\_nlp}.

First, we instantiate the NMs representing logical parts of the model: \texttt{TranslationDataLayer}, \texttt{TransformerEncoderNM}, \texttt{TransformerDecoderNM}, \texttt{TransformerLogSoftmaxNM}, and \texttt{PaddedSmoothedCrossEntropyLossNM}.


Then we construct the DAG of activation flow that looks like this:
\begin{lstlisting}[language=Python, basicstyle=\small]
src, src_mask, tgt, tgt_mask, labels, sent_ids = train_data_layer()
src_hiddens = encoder(input_ids=src, input_mask_src=src_mask)
tgt_hiddens = decoder(input_ids_tgt=tgt, hidden_states_src=src_hiddens,
                      input_mask_src=src_mask, input_mask_tgt=tgt_mask)
log_softmax = log_softmax(hidden_states=tgt_hiddens)
train_loss = loss(log_probs=log_softmax, target_ids=labels)
\end{lstlisting}
The code for training and callbacks is similar to the previous examples.

NeMo retains underlying framework's efficiency. In our experiment, this model achieves 29.2 BLEU / 28.5 SacreBLEU on newstest2014 after training for about 15 hours on WMT16 English-German using single machine with 8 GPUs.

\section{Conclusions and future work}\label{conclusions}
NeMo addresses many of the issues often encountered in developing DL applications by transferring best practices from software engineering. It operates with a higher level abstraction, the neural module, and introduces a neural type system capable of semantic checks. It also comes with collections of pre-built modules for conversational AI - \texttt{nemo\_asr} and \texttt{nemo\_nlp} to make building and re-using deep neural networks easier. 

We are working on expanding existing NeMo collections and adding new ones. Also, exploring the right design for a neural type system and the most useful levels of abstractions for modules is an ongoing research direction.

\bibliographystyle{plainnat}
\bibliography{main}

\begin{thebibliography}{20}
\providecommand{\natexlab}[1]{#1}
\providecommand{\url}[1]{\texttt{#1}}
\expandafter\ifx\csname urlstyle\endcsname\relax
  \providecommand{\doi}[1]{doi: #1}\else
  \providecommand{\doi}{doi: \begingroup \urlstyle{rm}\Url}\fi

\bibitem[PyT()]{PyTorchPipe}
Pytorchpipe.
\newblock \url{https://github.com/ibm/pytorchpipe}.
\newblock Accessed: 2019-08-19.

\bibitem[ape()]{apex}
A pytorch extension: Tools for easy mixed precision and distributed training in
  pytorch.
\newblock \url{https://github.com/NVIDIA/apex}.
\newblock Accessed: 2019-08-19.

\bibitem[ign()]{ignite}
Pytorch ignite.
\newblock \url{https://pytorch.org/ignite/}.
\newblock Accessed: 2019-08-19.

\bibitem[lig()]{lightning}
Pytorch lightning.
\newblock \url{https://github.com/williamFalcon/pytorch-lightning}.
\newblock Accessed: 2019-08-19.

\bibitem[lud()]{ludwig}
Ludwig.
\newblock \url{http://ludwig.ai}.
\newblock Accessed: 2019-08-19.

\bibitem[nam()]{namedtensor}
Tensor considered harmful.
\newblock \url{http://nlp.seas.harvard.edu/NamedTensor}.
\newblock Accessed: 2019-08-19.

\bibitem[ptr()]{ptransformers}
pytorch-transformers from huggingface.
\newblock \url{https://github.com/huggingface/pytorch-transformers}.
\newblock Accessed: 2019-08-19.

\bibitem[Abadi et~al.(2015)Abadi, Agarwal, Barham, Brevdo, Chen, Citro,
  Corrado, Davis, Dean, Devin, Ghemawat, Goodfellow, Harp, Irving, Isard, Jia,
  Jozefowicz, Kaiser, Kudlur, Levenberg, Man\'{e}, Monga, Moore, Murray, Olah,
  Schuster, Shlens, Steiner, Sutskever, Talwar, Tucker, Vanhoucke, Vasudevan,
  Vi\'{e}gas, Vinyals, Warden, Wattenberg, Wicke, Yu, and
  Zheng]{tensorflow2015-whitepaper}
Mart\'{\i}n Abadi, Ashish Agarwal, Paul Barham, Eugene Brevdo, Zhifeng Chen,
  Craig Citro, Greg~S. Corrado, Andy Davis, Jeffrey Dean, Matthieu Devin,
  Sanjay Ghemawat, Ian Goodfellow, Andrew Harp, Geoffrey Irving, Michael Isard,
  Yangqing Jia, Rafal Jozefowicz, Lukasz Kaiser, Manjunath Kudlur, Josh
  Levenberg, Dandelion Man\'{e}, Rajat Monga, Sherry Moore, Derek Murray, Chris
  Olah, Mike Schuster, Jonathon Shlens, Benoit Steiner, Ilya Sutskever, Kunal
  Talwar, Paul Tucker, Vincent Vanhoucke, Vijay Vasudevan, Fernanda Vi\'{e}gas,
  Oriol Vinyals, Pete Warden, Martin Wattenberg, Martin Wicke, Yuan Yu, and
  Xiaoqiang Zheng.
\newblock {TensorFlow}: Large-scale machine learning on heterogeneous systems,
  2015.
\newblock URL \url{https://www.tensorflow.org/}.
\newblock Software available from tensorflow.org.

\bibitem[{Britz} et~al.(2017){Britz}, {Goldie}, {Luong}, and {Le}]{Britz:2017}
Denny {Britz}, Anna {Goldie}, Thang {Luong}, and Quoc {Le}.
\newblock {Massive Exploration of Neural Machine Translation Architectures}.
\newblock \emph{ArXiv e-prints}, March 2017.

\bibitem[Chan et~al.(2016)Chan, Jaitly, Le, and Vinyals]{chan2016listen}
William Chan, Navdeep Jaitly, Quoc Le, and Oriol Vinyals.
\newblock Listen, attend and spell: A neural network for large vocabulary
  conversational speech recognition.
\newblock In \emph{2016 IEEE International Conference on Acoustics, Speech and
  Signal Processing (ICASSP)}, pages 4960--4964. IEEE, 2016.

\bibitem[Chollet et~al.(2015)]{chollet2015keras}
Fran\c{c}ois Chollet et~al.
\newblock Keras.
\newblock \url{https://keras.io}, 2015.

\bibitem[Klein et~al.(2017)Klein, Kim, Deng, Senellart, and
  Rush]{klein-etal-2017-opennmt}
Guillaume Klein, Yoon Kim, Yuntian Deng, Jean Senellart, and Alexander Rush.
\newblock {O}pen{NMT}: Open-source toolkit for neural machine translation.
\newblock In \emph{Proceedings of {ACL} 2017, System Demonstrations}, pages
  67--72, Vancouver, Canada, July 2017. Association for Computational
  Linguistics.
\newblock URL \url{https://www.aclweb.org/anthology/P17-4012}.

\bibitem[Kuchaiev et~al.(2018)Kuchaiev, Ginsburg, Gitman, Lavrukhin, Case, and
  Micikevicius]{kuchaiev2018}
Oleksii Kuchaiev, Boris Ginsburg, Igor Gitman, Vitaly Lavrukhin, Carl Case, and
  Paulius Micikevicius.
\newblock Openseq2seq: extensible toolkit for distributed and mixed precision
  training of sequence-to-sequence models.
\newblock \emph{arXiv e-prints arXiv:1805.10387}, 2018.

\bibitem[Li et~al.(2019)Li, Lavrukhin, Ginsburg, Leary, Kuchaiev, Cohen,
  Nguyen, and Gaddei]{li2019}
J.~Li, V.~Lavrukhin, B.~Ginsburg, R.~Leary, O.~Kuchaiev, J.~M. Cohen,
  H.~Nguyen, and R.~T. Gaddei.
\newblock Jasper: An end-to-end convolutional neural acoustic model.
\newblock \emph{arXiv e-prints arXiv:1904.03288}, 2019.

\bibitem[Micikevicius et~al.(2017)Micikevicius, Narang, Alben, Diamos, Elsen,
  Garc{\'{\i}}a, Ginsburg, Houston, Kuchaiev, Venkatesh, and
  Wu]{micikevicius2017}
Paulius Micikevicius, Sharan Narang, Jonah Alben, Gregory~F. Diamos, Erich
  Elsen, David Garc{\'{\i}}a, Boris Ginsburg, Michael Houston, Oleksii
  Kuchaiev, Ganesh Venkatesh, and Hao Wu.
\newblock Mixed precision training.
\newblock \emph{ICLR}, 2017.

\bibitem[Ott et~al.(2019)Ott, Edunov, Baevski, Fan, Gross, Ng, Grangier, and
  Auli]{ott2019fairseq}
Myle Ott, Sergey Edunov, Alexei Baevski, Angela Fan, Sam Gross, Nathan Ng,
  David Grangier, and Michael Auli.
\newblock fairseq: A fast, extensible toolkit for sequence modeling.
\newblock In \emph{Proceedings of NAACL-HLT 2019: Demonstrations}, 2019.

\bibitem[Paszke et~al.(2017)Paszke, Gross, Chintala, Chanan, Yang, DeVito, Lin,
  Desmaison, Antiga, and Lerer]{paszke2017automatic}
Adam Paszke, Sam Gross, Soumith Chintala, Gregory Chanan, Edward Yang, Zachary
  DeVito, Zeming Lin, Alban Desmaison, Luca Antiga, and Adam Lerer.
\newblock Automatic differentiation in pytorch.
\newblock 2017.

\bibitem[Reynolds et~al.(2017)Reynolds, Barth-Maron, Besse, de~Las~Casas,
  Fidjeland, Green, Puigdom{\`e}nech, Racani{\`e}re, Rae, and
  Viola]{sonnetblog}
Malcolm Reynolds, Gabriel Barth-Maron, Frederic Besse, Diego de~Las~Casas,
  Andreas Fidjeland, Tim Green, Adri{\`a} Puigdom{\`e}nech, S{\'e}bastien
  Racani{\`e}re, Jack Rae, and Fabio Viola.
\newblock {Open sourcing Sonnet - a new library for constructing neural
  networks}.
\newblock \url{https://deepmind.com/blog/open-sourcing-sonnet/}, 2017.

\bibitem[Vaswani et~al.(2017)Vaswani, Shazeer, Parmar, Uszkoreit, Jones, Gomez,
  Kaiser, and Polosukhin]{vaswani2017}
Ashish Vaswani, Noam Shazeer, Niki Parmar, Jakob Uszkoreit, Llion Jones,
  Aidan~N. Gomez, Lukasz Kaiser, and Illia Polosukhin.
\newblock Attention is all you need.
\newblock \emph{arXiv preprint arXiv: 1706.03762}, 2017.

\bibitem[Vaswani et~al.(2018)Vaswani, Bengio, Brevdo, Chollet, Gomez, Gouws,
  Jones, Kaiser, Kalchbrenner, Parmar, Sepassi, Shazeer, and
  Uszkoreit]{tensor2tensor}
Ashish Vaswani, Samy Bengio, Eugene Brevdo, Francois Chollet, Aidan~N. Gomez,
  Stephan Gouws, Llion Jones, \L{}ukasz Kaiser, Nal Kalchbrenner, Niki Parmar,
  Ryan Sepassi, Noam Shazeer, and Jakob Uszkoreit.
\newblock Tensor2tensor for neural machine translation.
\newblock \emph{CoRR}, abs/1803.07416, 2018.
\newblock URL \url{http://arxiv.org/abs/1803.07416}.

\end{thebibliography}

\end{document}